\documentclass[11pt,a4paper]{article}

\usepackage[hyperref]{acl2017}
\usepackage{times}
\usepackage{latexsym}
\usepackage{subcaption}
\usepackage{tabularx}

\usepackage[hyphenbreaks]{breakurl}

\usepackage{tikz}
\usetikzlibrary{matrix,chains,positioning,decorations.pathreplacing,arrows}

\newcommand{\com}[1]{}

\aclfinalcopy

\def\aclpaperid{18}

 \author{Yftah Ziser \and Roi Reichart \\
 Faculty of Industrial Engineering and Management, Technion, IIT\\
        syftah@campus.technion.ac.il, roiri@ie.technion.ac.il}

\date{}

\title{Neural Structural Correspondence Learning for Domain Adaptation}

\date{}

\begin{document}
\maketitle
\begin{abstract}
  

     We introduce a neural network model that marries together ideas 
     from two prominent strands of research on domain adaptation through representation learning:  structural correspondence learning (\textit{SCL}, \cite{Blitzer:06}) and autoencoder neural 
     networks (NNs). 
Our model is a three-layer NN that learns to encode the 
non-pivot features of an input example into a low-dimensional representation, so that 
the existence of pivot features (features that are prominent in both domains 
and convey useful information for the NLP task) in the example can be decoded from that representation.
The low-dimensional representation is then employed in a learning algorithm for the task.
Moreover, we show how to inject pre-trained word embeddings into our model in order to improve generalization
across examples with similar pivot features. 
We experiment with the task of cross-domain sentiment classification
on 16 domain pairs and show substantial improvements over strong baselines.\footnote{Our code is at: https://github.com/yftah89/Neural-SCL-Domain-Adaptation.}


\end{abstract}
\section{Introduction}

Many state-of-the-art algorithms for Natural Language Processing (NLP) tasks require labeled data. 
Unfortunately, annotating sufficient amounts of such data is often costly and labor intensive. Consequently, for many 
NLP applications even resource-rich languages like English have labeled data in only a handful of domains. 

Domain adaptation \cite{Daume:07,BenDavid:10}, training an algorithm on labeled data taken 
from one domain so that it can perform properly on data from other domains, is therefore recognized as a 
fundamental challenge in NLP. Indeed, over the last decade domain adaptation methods have been proposed for tasks such 
as sentiment classification \cite{Bollegala:11a}, POS tagging \cite{Schnabel:13}, syntactic parsing 
\cite{Reichart:07,Mcclosky:10,Rush:12} and relation extraction \cite{Jiang:07,Bollegala:11b}, if to name just a handful 
of applications and works.

Leading recent approaches to domain adaptation in NLP are based on Neural Networks (NNs), and particularly on autoencoders \cite{Glorot:11,Chen:12}. These models are believed to extract features that are robust to cross-domain variations. However, while excelling  
on benchmark domain adaptation tasks such as cross-domain product sentiment classification \cite{Blitzer:07}, 
the reasons to this success are not entirely understood.

In the pre-NN era, a prominent approach to domain adaptation in NLP, and particularly in sentiment classification, has been
structural correspondence learning (\textit{SCL)} \cite{Blitzer:06,Blitzer:07}. Following the auxiliary problems 
approach to semi-supervised learning \cite{Ando:05}, this method identifies correspondences 
among features from different domains by modeling their correlations with \textit{pivot} features: features 
that are frequent in both domains and are important for the NLP task. Non-pivot features from different domains
which are correlated with many of the same pivot features are assumed to correspond, providing a bridge 
between the domains. Elegant and well motivated 
as it may be, SCL does not form the 
state-of-the-art since the
neural approaches took over. 

In this paper we marry these approaches, proposing NN models 
inspired by ideas from both. Particularly, our basic model  
receives the non-pivot features of an input example, encodes them into a hidden layer and then, instead of decoding the input layer as an 
autoencoder would do, it aims to decode the pivot features. 
Our more advanced model is identical to the basic one except that the decoding matrix is not learned but is 
rather replaced with a fixed matrix consisting of pre-trained embeddings of the pivot features. Under this model the 
probability of the $i$-th pivot feature to appear in an example is a (non-linear) function of the dot product of
the feature's embedding vector and the network's hidden layer vector.
As explained in Section \ref{sec:model}, this approach encourages the model to learn similar hidden layers 
for documents that have different pivot features as long as these features have similar meaning. In sentiment classification, for example,  
although one positive review may use the unigram pivot feature {\it excellent} while another positive review uses the pivot {\it great}, as long as the embeddings of pivot features with similar meaning are similar (as expected from high quality embeddings) the hidden layers learned for both documents are biased to be similar.

We experiment with the task of cross-domain product sentiment classification of \cite{Blitzer:07}, consisting of 4 domains (12 domain pairs) and further add an additional target domain, consisting of sentences extracted from social media blogs (total of 16 domain pairs).  For pivot feature embedding in our advanced model, we employ the word2vec algorithm \cite{Mikolov:13}. 
Our models substantially outperform strong baselines: the SCL algorithm, the marginalized stacked denoising autoencoder (MSDA) model \cite{Chen:12} and the MSDA-DAN model \cite{Ganin:16} that combines the power of MSDA with a domain adversarial network (DAN). 
\com{
On the 12 product domain pairs of \cite{Blitzer:07} our advanced model achieves an averaged accuracy of 78.08\%, compared to 77.04\% of our basic model, 
75.92\% of MSDA, 74.31\% of SCL and 73.05\% of a non-adapted classifier. When adapting from the four product review domains to the blog domain our average gains are even larger: 76.9\% average accuracy for the advanced model, 71.8\% for the basic model, 71.7\% for MSDA, 70.5\% for SCL and only 65.2\% for a non-adapted classifier\footnote{We will release our code upon paper acceptance.}
}

\section{Background and Contribution}
\label{sec:prev}

Domain adaptation is a fundamental, long standing problem in NLP  (e.g. \cite{Roark:03,Chelba:04,Daume:06}). 
The challenge stems from the fact that data in the source and the target domains 
are often distributed differently, making it hard for a model trained in the source domain
to make valuable predictions in the target domain. 

Domain adaptation has various setups, 
differing with respect to the amounts of labeled and unlabeled data available in the source and 
target domains. The setup we address, commonly referred to as unsupervised domain adaptation
is where both domains have ample unlabeled data, but only the source domain has labeled training data. 

There are several approaches to domain adaptation in the machine learning literature, including 
instance reweighting \cite{Huang:07,Mansour:09}, sub-sampling from both
domains \cite {Chen:11} and learning joint target and source feature representations 
\cite{Blitzer:06,Daume:07,Xue:08,Glorot:11,Chen:12}. 

Here, we discuss works that, like us, take the representation learning path. Most works under this approach follow a two steps protocol: First, the representation learning method (be it SCL, an autoencoder network, our proposed network model or any other model) is trained on unlabeled data from both the source and the target domains; Then, a classifier for the supervised task (e.g. sentiment classification) is trained in the source domain and this trained classifier is applied to test examples from the target domain. Each input example of the task classifier, at both training and test, is first run through the representation model of the first step and the induced representation is fed to the classifier. Recently, end-to-end models that jointly learn to represent the data and to perform the classification task have also been proposed. We compare our models to one such method (MSDA-DAN, \cite{Ganin:16}). 

Below, we first discuss two prominent ideas in feature representation learning: pivot features 
and autoencoder neural networks. 
We then summarize our contribution in light of these approaches.

\paragraph {Pivot and Non-Pivot Features}
The definitions of this approach are given in 
\newcite{Blitzer:06,Blitzer:07}, where SCL is presented in the context of POS tagging and sentiment classification, respectively. 
Fundamentally, the method divides the shared feature space of both the source and the target domains 
to the set of pivot features that are frequent in both domains and are prominent in the NLP task, 
and a complementary set of non-pivot features. In this section we abstract away from the 
actual feature space and its division to pivot and non-pivot subsets. In Section \ref{sec:experiments} we 
discuss this issue in the context of  sentiment classification.

For representation learning, SCL employs the pivot features in order to learn mappings from the original feature space of both domains to a shared, low-dimensional, 
real-valued feature space. This is done by training classifiers whose input consists of the non-pivot features of an input example and their binary classification task (the auxiliary task) is predicting, every classifier for one pivot feature, whether the pivot associated with the classifier appears in the input example or not. 
These classifiers are trained on unlabeled data from both the target and the source domains: the training supervision naturally occurs in the data, no human annotation is required.
The matrix consisting of the weight vectors of these classifiers is then post-processed with singular 
value decomposition (SVD), to facilitate final compact representations. 
The SVD derived matrix serves as a transformation matrix which maps feature vectors in 
the original space into a low-dimensional real-valued feature space.

\com{In the second step of SCL, a classifier for the original learning task is trained on the source domain, where labeled 
data exists for this task. The input feature representation in this classifier is an augmentation of the original 
feature vector and the one derived from the above transformation. This classifier is finally employed to the 
target domain, where no labeled data exists.}

Numerous works have employed the SCL method in particular and the concept of pivot features 
for domain adaptation in general. 
A prominent method is spectral feature
alignment (SFA, \cite{Pan:10}). This method aims to align domain-specific (non-pivot) features from
different domains into unified clusters, with the help of domain-independent (pivot)
features as a bridge. 

Recently, \newcite{Gouws:12} and \newcite{Bollegala:15} implemented ideas related to those described here within an NN for cross-domain sentiment classification. 
For example, the latter work trained a word embedding model so that for every document, regardless of its domain, pivots are good predictors of non-pivots, and the pivots' embeddings are similar across domains.
\newcite{Yu:16} presented a convolutional NN that learns sentence embeddings using two auxiliary tasks (whether the sentence contains a positive or a negative domain independent sentiment word), purposely avoiding prediction with respect to a large set of pivot features. 
In contrast to these works our model can learn useful cross-domain representations for any type of input example and in our cross-domain sentiment classification experiments it learns document level embeddings. 
That is, unlike \newcite{Bollegala:15} we do not learn word embeddings and unlike \newcite{Yu:16} we are not restricted to input sentences.



\noindent \textbf {Autoencoder NNs}
An autoencoder is comprised of an encoder function $h$
and a decoder function $g$, typically with the dimension of $h$ smaller than that of its argument. 
The reconstruction of an input $x$ is given by $r(x) = g(h(x))$.
Autoencoders are typically trained to minimize a reconstruction error $loss(x,r(x))$.
Example loss functions are the squared error, the Kullback-Leibler (KL) divergence
and the cross entropy of elements of $x$ and elements of $r(x)$. 
The last two loss functions are appropriate options when 
the elements of $x$ or $r(x)$ can be interpreted as probabilities of a discrete event.
In Section \ref{sec:model} we get back to this point when defining the cross-entropy loss function 
of our model. 
Once an autoencoder has been trained, one can stack another autoencoder on top of it, by training a second
model which sees the output of the first as its training data \cite{Bengio:07}.
The parameters of the stack of autoencoders describe multiple representation levels for $x$ and can feed a classifier, to facilitate domain adaptation.

Recent prominent models for domain adaptation for sentiment classification are based on a variant of the 
autoencoder called Stacked Denoising Autoencoders (SDA, \cite{Vincent:08}). 
In a denoising autoencoder (DEA) the input vector $x$ is stochastically
corrupted into a vector $\tilde{x}$, and the model is trained to minimize a denoising reconstruction
error $loss(x,r(\tilde{x}))$. SDA for cross-domain sentiment classification was implemented by \newcite{Glorot:11}. 
Later, \newcite{Chen:12} proposed the marginalized SDA (MSDA) model that is more computationally efficient and scalable 
to high-dimensional feature spaces than SDA. 

Marginalization of denoising autoencoders has gained interest since MSDA was presented. \newcite{Yang:14} showed how to improve efficiency further by exploiting noising functions designed for structured feature
spaces, which are common in NLP. More recently, \newcite{Clinchant:16} proposed an unsupervised regularization method for MSDA based on the work of \newcite{Ganin:15} and \newcite{Ganin:16}. 

There is a recent interest in models based on variational autoencoders \cite{Kingma:13,Rezende:14}, for example the variational fair autoencoder model \cite{Louizos:16}, for domain adaptation. However, these models are still not competitive with MSDA on the tasks we consider here.


\paragraph{Our Contribution} 
We propose an approach that marries the above lines of work. 
Our model is similar in structure to an 
autoencoder. However, instead of reconstructing the input $x$ from the hidden layer $h(x)$, 
its reconstruction function $r$ receives a low dimensional representation of 
the non-pivot features of the input 
($h(x^{np})$, where $x^{np}$ is the non-pivot representation of $x$ (Section \ref{sec:model})) 
and predicts whether each of the pivot features appears in this example or not.
 As far as we know, we are the first to exploit the mutual strengths of pivot-based methods and autoencoders for domain adaptation.

\com{
\paragraph{Our Contribution} this paper we propose an approach that marries the above two lines of work. 
It does so by providing an NN model, inspired by  autoencoder networks, to the idea of domain adaptation through exploitation of the relationship between pivot and non-pivot features.

Our model is similar in structure to an 
autoencoder NN. However, instead of reconstructing the input $x$ from the hidden layer $h(x)$, 
its reconstruction function $r$ receives a low dimensional representation of 
the non-pivot features as an input 
($h(x^{np})$ where $x^{np}$ is the non-pivot representation of $x$, see Section \ref{sec:model}) 
and predicts whether each of the pivot features appears in this example or not.\footnote{Following SCL, we feed the task classifier with a concatenation of the original and the learned representations.} 

To the best of our knowledge, we are the first 
to exploit the mutual strengths of pivot-based methods and autoencoder NNs for domain adaptation.
}

\section{Neural SCL Models}
\label{sec:model}

We propose two models:
the basic \textit{Autoencoder SCL (AE-SCL, \ref{sec:basic}))}, that directly integrates ideas from autoencoders and SCL, and the elaborated \textit{Autoencoder SCL with Similarity Regularization (AE-SCL-SR, \ref{sec:w2v})}, where pre-trained word embeddings are integrated into the basic model.


\subsection{Definitions}
\label{sec:definitions}

We denote the feature set in our problem with $f$, the subset of pivot features with 
$f_p \subseteq \{1, \ldots, |f|\}$ and the subset of non-pivot features with $f_{np} \subseteq \{1, \ldots, |f|\}$ 
such that $f_p \cup f_{np} = \{1, \ldots, |f|\} $ and $f_p \cap f_{np} = \emptyset$. 
We further denote the feature representation of an input example $X$ with $x$. Following this notation, 
the vector of pivot features of $X$ is denoted with $x^{p}$ while the vector of non-pivot features
is denoted with $x^{np}$.  

In order to learn a robust and compact feature representation for $X$ we will aim to learn a non-linear
prediction function from $x^{np}$ to $x^{p}$. As discussed in Section \ref{sec:experiments} the task we experiment with is cross-domain sentiment classification. Following previous work 
(e.g. \cite{Blitzer:06,Blitzer:07,Chen:12} our feature representation 
consists of binary indicators for the occurrence of word unigrams and bigrams in the represented 
document. In what follows we hence assume that the feature representation $x$ of an example $X$ is a 
binary vector, and hence so are $x^{p}$ and $x^{np}$.

\subsection{Autoencoder SCL (AE-SCL)}
\label{sec:basic}

In order to solve the prediction problem, we present an NN architecture inspired by autoencoders (Figure 1). Given an input example $X$ with a feature representation $x$, our fundamental idea is to start from a non-pivot 
feature representation, $x^{np}$, encode $x^{np}$ into an intermediate representation $h_{w^h}(x^{np})$, and, finally, predict with a function $r_{w^r}(h_{w^h}(x^{np}))$ the occurrences of pivot features, $x^{p}$, in the example.

As is standard in NN modeling, we introduce non-linearity to the model through a 
non-linear activation function denoted with $\sigma$ (the sigmoid function in our models). Consequently we get:
$h_{w^h}(x^{np}) = \sigma(w^hx^{np})$ and $r_{w^r}(h_{w^h}(x^{np})) = \sigma(w^rh_{w^h}(x^{np}))$.
In what follows we denote the output of the model with $o = r_{w^r}(h_{w^h}(x^{np}))$. 

Since the sigmoid function outputs values in the $[0,1]$ interval, 
$o$ can be interpreted as a vector of probabilities with the $i$-th coordinate 
reflecting the probability of the $i$-th pivot feature to appear in the input example. 
Cross-entropy is hence a natural loss function to jointly reason about all pivots:
\[L(o,x^p) = \frac{1}{|f_p|}\sum_{i=1}^{|f_p|} {x^p}_i \cdot log (o_i) + (1-{x^p}_i) \cdot log(1-o_i)\]

As $x^p$ is a binary vector, for each pivot feature, ${x^p}_i$, only one of the two members of the sum that take this feature into account gets a non-zero value. 
The higher the probability of the correct event is (whether or not ${x^p}_i$ appears in the input example), the lower is the loss.

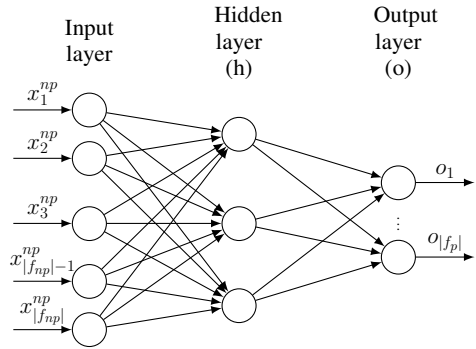
\begin{figure}[h!]
	\centering

\begin{tikzpicture}[scale=0.5, every node/.style={scale=0.5},
plain/.style={
    draw=none,
    fill=none,
},
net/.style={
    matrix of nodes,
    nodes={
        draw,
        circle,
        inner sep=9pt
    },
    nodes in empty cells,
    column sep=0.5cm,
    row sep=-4pt
},
>=latex
]
\matrix[net] (mat)
{
    |[plain]| \parbox{1.3cm}{\LARGE\centering Input\\layer} & |[plain]| \parbox{1.3cm}{\LARGE\centering Hidden\\layer (h)} & |[plain]| \parbox{1.3cm}{\centering \LARGE Output\\layer (o)} \\
    
    & |[plain]| \\
    |[plain]| &  \\
    & |[plain]|   \\
    |[plain]| & |[plain]|& \\
    & &|[plain]| \parbox{0.5cm}{\centering $\vdots$} \\
     |[plain]|& |[plain]| & \\
    & |[plain]| \\
    |[plain]| & \\
    & |[plain]| \\
};
\foreach \ai [count=\mi ]in {2,4,...,6}
\draw[<-] (mat-\ai-1) -- node[above] {\LARGE $x^{np}_\mi$ } +(-2cm,0);
\foreach \ai in {2,4,...,10}
{\foreach \aii in {3,6,9}
    \draw[->] (mat-\ai-1) -- (mat-\aii-2);
}
\foreach \ai in {3,6,9}{
\draw[->] (mat-\ai-2) -- (mat-5-3);
\draw[->] (mat-\ai-2) -- (mat-7-3);
}
\draw[->] (mat-5-3) -- node[above] {\LARGE $o_1$} +(2cm,0);
\draw[->] (mat-7-3) -- node[above] {\LARGE $o_{|f_p|}$} +(2cm,0);
\draw[opacity=0.0, text opacity=1] (mat-6-1) -- (mat-8-1) node[midway] {$\vdots$};
\draw[opacity=0.0, text opacity=1] (mat-6-2) -- (mat-9-2) node[midway] {$\vdots$};

\draw[opacity=0.0, text opacity=1] (mat-1-1) -- (mat-1-2) node[midway] {\LARGE$w_h$};
\draw[opacity=0.0, text opacity=1] (mat-1-2) -- (mat-1-3) node[midway] {\LARGE$w_r$};
\draw[<-] (mat-10-1) -- node[above] {\LARGE$x^{np}_{|f_{np}|}$ } +(-2cm,0);
\draw[<-] (mat-8-1) -- node[above] {\LARGE$x^{np}_{|f_{np}|-1}$ } +(-2cm,0);

\end{tikzpicture}
\caption{A Sketch of the AE-SCL and AE-SCL-SR models. 
While in AE-SCL both the encoding matrix $w^h$ and the reconstruction matrix $w^r$ are optimized, 
in AE-SCL-SR $w^{r}$ is pre-trained by a word embedding model. 
See full details in text.
}
\end{figure}

\subsection{Autoencoder SCL with Similarity Regularization (AE-SCL-SR)}
\label{sec:w2v}

An important observation of \newcite{Blitzer:07}, is that some pivot features are similar to each other to the level that they indicate the same information with respect to the classification task. 
For example, in sentiment classification with word unigram features, the words (unigrams) \textit{great} 
and \textit{excellent} are likely to serve as pivot features, as the meaning of each of them is preserved 
across domains. 
At the same time, both features convey very similar (positive) sentiment information to the level that a sentiment classifier should treat them as equals.

The AE-SCL-SR model is based on two crucial observations. First, in many NLP tasks the pivot features can be pre-embeded into a vector space where pivots with similar meaning have similar vectors. Second, the set ${f_p}^{X_i}$  of pivot features that appear in an example 
$X_i$ is typically much smaller than the set $\hat{{f_p}^{X_i}}$ of pivot features that do not appear in it. 
Hence, if the pivot features of $X_1$ and $X_2$ convey the same information about the NLP task (e.g. that the sentiment of both $X_1$ and $X_2$ is positive), 
then even if ${f_p}^{X_1}$  and ${f_p}^{X_2}$ are not identical, the intersection between the larger sets 
$\hat{{f_p}^{X_1}}$ and $\hat{{f_p}^{X_2}}$ is typically much larger than the symmetric difference between ${f_p}^{X_1}$  and ${f_p}^{X_2}$.


For instance, consider two examples, $X_1$ with the single pivot feature $f_1 = great$, and $X_2$, with the single pivot feature $f_2 = excellent$. 
Crucially, even though $X_1$ and $X_2$ differ with respect to the existence of 
$f_1$ and $f_2$, due to the similar meaning of these pivot features, we expect both $X_1$ and $X_2$ not to contain 
many other pivot features, such as \textit{terrible, awful} and \textit{mediocre}, whose meanings conflict with that of $f_1$ and $f_2$. 

To exploit these observations, in AE-SCL-SR the reconstruction matrix $w^{r}$ is 
pre-trained with a word embedding model and is kept fixed during the training and prediction phases of the neural 
network. Particularly, the $i$-th row of $w^{r}$ is set to be the vector representation of the $i$-th pivot feature as 
learned by the word embedding model.
Except from this change, the AE-SCL-SR model is identical to the 
AE-SCL model described above.


Now, denoting the encoding layer for $X_1$ with $h_1$ and the encoding layer for $X_2$ with $h_2$, we expect both 
$\sigma(w^{r}_{\vec{k_i}} \cdot h_1)$ and $\sigma(w^{r}_{\vec{k_i}} \cdot h_2)$ to get low values (i.e. values close to 0), 
for those $k_i$ \textit{conflicting pivot features}: pivots whose meanings conflict with that of ${f_p}^{X_1}$  and ${f_p}^{X_2}$. 
By fixing the representations of similar conflicting features to similar vectors, AE-SCL-SR provides a strong bias for $h_1$ and $h_2$ to be similar, as its only way to bias the predictions with respect to these features to be low is by pushing $h_1$ and $h_2$ to be similar. Consequently, under AE-SCL-SR the vectors that encode the non-pivot features of documents with similar pivot features are biased to be similar to each other. 
As mentioned in Section \ref{sec:experiments} 
the vector $\tilde{h} = \sigma^{-1}(h)$ forms the feature representation that is fed to the sentiment classifier
to facilitate domain adaptation.
By definition, when $h_1$ and $h_2$ are similar so are their $\tilde{h_1}$ and $\tilde{h_2}$ counterparts.


\com{
We define the vector $\vec{x}_{R_{1x|\mathcal{N}|}}$ in the following way:
\begin{equation}
\vec{x}_{R_{i}} = 
\begin{cases}
1, & \text{if}\ \mathcal{N}_i \in x_r\\
0, & \text{otherwise}
\end{cases}
\end{equation}
Similarly, we define the vector $\vec{y}_{R_{1x|\mathcal{P}|}}$ in the following way:
\begin{equation}
\vec{y}_{R_{i}} = 
\begin{cases}
1, & \text{if}\ \mathcal{P}_i \in x_r\\
0, & \text{otherwise}
\end{cases}
\end{equation}
We can now look at any review as a prediction problem in which for the binary vector $\vec{x}_R$ we would like to predict the corresponding binary vector $\vec{y}_R$.
}

\section{Experiments} 

\label{sec:experiments}

In this section we describe our experiments. To facilitate clarity, some  details are not given here and instead are provided in the appendices.

\noindent \textbf{Cross-domain Sentiment Classification}
To demonstrate the power of our models for domain adaptation we experiment with the task of cross-domain sentiment 
classification \cite{Blitzer:07}. The data for this task consist of Amazon product reviews from 
four product domains: Books (B), DVDs (D), 
Electronic items (E) and Kitchen appliances (K).
For each domain 2000 labeled reviews are provided: 1000 are classified as positive and 1000 as negative, 
and these are augmented with unlabeled reviews: 6000 (B), 
34741 (D), 13153 (E) and 16785 (K).

We also consider an additional target domain, denoted with \textit{Blog}: the University of Michigan sentence level sentiment dataset, consisting of sentences taken from social media blogs.\footnote{https://inclass.kaggle.com/c/si650winter11}
The dataset for the original task consists of a labeled training set (3995 positive and 3091 negative) and a 33052 sentences test set for which sentiment labels are not provided. We hence used the original test set as our target domain unlabeled set and the original training set as our target domain test set.


\paragraph {Baselines}
Cross-domain sentiment classification has been 
studied in a large number of papers. However, 
the difference in preprocessing methods, dataset splits to train/dev/test subsets and the different sentiment 
classifiers make it hard to directly compare between the numbers reported in past.

We hence compare our models to three strong baselines, running all models 
under the same conditions. 
We aim to select baselines that represent the state-of-the-art in cross-domain sentiment classification in general, and in the two lines of work we focus at: pivot based and autoencoder based representation learning, in particular.

The first baseline is SCL with pivot features selected using the mutual 
information criterion (SCL-MI, \cite{Blitzer:07}). This is the SCL method where pivot features are frequent in the unlabeled data of both the source and the target 
domains, and among those features are the ones with the highest mutual 
information with the task (sentiment) label in the source domain labeled data.
We implemented this method. In our implementation unigrams and bigrams should appear at least 10 times in both domains to be considered frequent. For non-pivot features we consider unigrams and bigrams that appear at least 10 times in their domain. The same pivot and non-pivot selection criteria are employed for our AE-SCL and AE-SCL-SR models.

Among autoencoder models, SDA has shown by \newcite{Glorot:11} to outperform SFA and SCL on cross-domain sentiment classification and later on \newcite{Chen:12} demonstrated superior performance for MSDA over SDA and SCL on the same task. Our second baseline is hence the MSDA method \cite{Chen:12}, with code taken from the authors' 
web page.\footnote{\url{http://www.cse.wustl.edu/~mchen}}

To consider a regularization scheme on top of MSDA representations we also experiment with the MSDA-DAN model \cite{Ganin:16} which employs a domain adversarial network (DAN) with the MSDA vectors as input. In \newcite{Ganin:16} MSDA-DAN has shown to substantially outperform the DAN model when DAN is randomly initialized. The DAN code is taken from the authors' repository. \footnote{\url{ https://github.com/GRAAL-Research/domain_adversarial_neural_network}} 

For reference we compare to the No-DA case where the sentiment classifier is trained in the source domain and applied to the target domain without adaptation. The sentiment classifier we employ, in this case as well as with our methods and with the SCL-MI and MSDA baselines, is a standard logistic regression classifier.\footnote{\url{http://scikit-learn.org/stable/}} \footnote{We tried to compare to \cite{Bollegala:15} but failed to replicate their results despite
personal communication with the authors.}

\paragraph {Experimental Protocol}
Following the unsupervised domain adaptation setup (Section~\ref{sec:prev}), we have access to unlabeled data from both the source and the target domains, which we use to train the representation learning models. However, only the source domain has labeled training data for sentiment classification. The original feature set we start from consists of word unigrams and bigrams.

All methods (baselines and ours), except from MSDA-DAN, follow a two-step protocol at both training and test time. In the first step, the input example is run through the representation model which generates a new feature vector for this example. Then, in the second step, this vector is concatenated with the original feature vector of the example and the resulting vector is fed into the sentiment classifier (this concatenation is a standard convention in the baseline methods).

For MSDA-DAN all the above holds, except from one exception. MSDA-DAN gets an input representation that consists of a concatenation of the original and the MSDA-induced feature sets. As this is an end-to-end model that predicts the sentiment class jointly with the new feature representation, we do not employ any additional sentiment classifier. As in the other models, MSDA-DAN utilizes source domain labeled data as well as unlabeled data from both the source and the target domains at training time. 


We experiment with a 5-fold cross-validation 
on the source domain \cite{Blitzer:07}: 
1600 reviews for training  and 400 reviews for development. 
The test set for each target domain of \newcite{Blitzer:07} consists of all 2000 labeled reviews of that domain, and for the Blog domain it consists of the 7086 labeled sentences provided with the task dataset.
In all five folds half of the training examples and half of the development examples are randomly selected 
from the positive reviews and the other halves from the negative reviews.
We report average results across these five folds, employing the same folds for all models.

\begin{table*}[t!]
\small
	\centering
	\begin{tabular}{|l|l|l|l|l|l|l|}
		\hline
Model$|$Source$\rightarrow$Target	& \textbf{D$\rightarrow$B}  & \textbf{E$\rightarrow$B}   & \textbf{K$\rightarrow$B}   & \textbf{B$\rightarrow$D}   & \textbf{E$\rightarrow$D}  & \textbf{K$\rightarrow$D}  \\ \hline
		\textbf{AE-SCL-SR} & $\mathbf{0.773^{*+\diamond}}$ & 0.7115          & $0.730^{*+}$          & $\mathbf{0.811^{+\diamond}}$ & $\mathbf{0.745^{*+\diamond}}$ & $\mathbf{0.763^{*+\diamond}}$ \\ \hline
                \textbf{AE-SCL}    & $0.758^{\odot}$         & 0.701           & $\mathbf{0.742^{\ddagger\odot}}$ & 0.794           & $0.732^{\ddagger}$          & $0.743^{\ddagger\odot}$          \\ \hline
		\textbf{MSDA}     & 0.761         & \textbf{0.719} & 0.7             & 0.783           & 0.71           & 0.714          \\ \hline	
         \textbf{MSDA-DAN}     & 0.75         & 0.71 & 0.712             & 0.797           & 0.731           & 0.738          \\ \hline	
		\textbf{SCL-MI}      & 0.732          & 0.685          & 0.693          & 0.788           & 0.704          & 0.722         \\ \hline
		\textbf{No-DA}    & 0.736          & 0.679           & 0.677          & 0.76            & 0.692         & 0.702          \\ \hline
	\end{tabular}
	\bigskip
\\
	\begin{tabular}{|l|l|l|l|l|l|l|l|}
		\hline
Mod.$|$So.$\rightarrow$Tar.& \textbf{B$\rightarrow$E}   & \textbf{D$\rightarrow$E}  & \textbf{K$\rightarrow$E} & \textbf{B$\rightarrow$K}   & \textbf{D$\rightarrow$K}   & \textbf{E$\rightarrow$K} & \textbf{Test-All}   \\ \hline
		\textbf{AE-SCL-SR}  & $\mathbf{0.768^{*+\diamond}}$ & $\mathbf{0.781^{*+\diamond}}$ & $\mathbf{0.84^{*+\diamond}}$ & $\mathbf{0.801^{*+\diamond}}$ & $\mathbf{0.803^{+\diamond}}$ & $0.846$  & $\mathbf {0.781^{*+\diamond}}$ \\ \hline
		\textbf{AE-SCL}    & 0.744          & $0.763^{\ddagger\odot}$          & $0.828^{\odot}$        & $0.795^{\ddagger\odot}$  & $0.8^{\ddagger\odot}$  & 0.848 & $0.770^{\ddagger}$ \\ \hline
		\textbf{MSDA}     & 0.746          & 0.75           & 0.824         & 0.788           & 0.774           & 0.845           & 0.759  \\ \hline
        \textbf{MSDA-DAN}     & 0.747          & 0.745           & 0.821         & 0.754           & 0.776           & \textbf{0.85}           & 0.761  \\ \hline
		\textbf{SCL-MI}      & 0.719          & 0.715          & 0.822         & 0.772          & 0.74          & 0.829           & 0.743  \\ \hline
		\textbf{No-DA}    & 0.7             & 0.709          & 0.816         & 0.74            & 0.732           & 0.824           & 0.731  \\ \hline
	\end{tabular}
	\bigskip
\\
    \begin{tabular}{|l|l|l|l|l|l|l|l|}
		\hline
        Mod.$|$So.$\rightarrow$Tar.        & \textbf{B$\rightarrow$Blog} & \textbf{D$\rightarrow$Blog}& \textbf{E$\rightarrow$Blog}& \textbf{K$\rightarrow$Blog}& \textbf{Test-All}\\ \hline
        \textbf{AE-SCL-SR}  & $\mathbf{0.705^{*+}}$ & $\mathbf{0.793^{+\diamond}}$ & $0.703^{*+\diamond}$ & $\mathbf{0.841^{*+\diamond}}$ & $\mathbf{0.769^{*+\diamond}}$ \\ \hline
        \textbf{AE-SCL}  & 0.691 & $0.787^{\ddagger\odot}$ & $0.645^{\odot}$ & $0.747^{\odot}$ & 0.718\\ \hline
        \textbf{MSDA}  & 0.698 & 0.775 & 0.646 & 0.75 & 0.717 \\ \hline
        \textbf{MSDA-DAN}  & 0.694 & 0.737 & \textbf{0.764} & 0.672 & 0.716 \\ \hline
        \textbf{SCL-MI}  & 0.687 & 0.767 & 0.662 & 0.704 & 0.705 \\ \hline
        \textbf{NO-DA}  & 0.627 & 0.747 & 0.620 & 0.616 & 0.652 \\  \hline
        \end{tabular}
\caption{
Sentiment classification accuracy for the \newcite{Blitzer:07} task (top tables), and 
for adaptation from the Blitzer's product review domains to the Blog domain (bottom table).
Test-All presents average results across setups. Statistical significance (with the McNemar
paired test for labeling disagreements \cite{Gillick:89,Blitzer:06}, $p < 0.05$) is denoted with:
$*$ (AE-SCL-SR vs. AE-SCL), $+$ (AE-SCL-SR vs. MSDA), $\diamond$ (AE-SCL-SR vs. MSDA-DAN), $\ddagger$ (AE-SCL vs. MSDA) and 
$\odot$ (AE-SCL vs. MSDA-DAN).
All the differences between any model and No-DA are statistically significant.
}
\label{tab:results}
\end{table*}

\paragraph{Hyper-parameter Tuning}

The details of the hyper-parameter tuning process for all  models (including data splits to training, development and test sets) are described in the appendices. Here we provide a summary.

\noindent \textit{AE-SCL and AE-SCL-SR:} For the stochastic gradient descent (SGD) training algorithm we set the learning rate to 0.1, momentum to 0.9 and weight-decay regularization to $10^{-5}$. The number of pivots was chosen among $\{100, 200, \ldots  ,500\}$
and the dimensionality of $h$ among $\{100, 300, 500\}$. For the features induced by these models we take their $w^{h}x^{np}$ vector.
For AE-SCL-SR, embeddings for the unigram and bigram features were  learned with word2vec \cite{Mikolov:13}. Details about the software  
and the way we learn bigram representations are in the appendices.

\noindent \textit{Baselines:} For SCL-MI, following \cite{Blitzer:07} we tuned the number of pivot features between 500 and 1000 and the SVD dimensions among 50,100 and 150. For MSDA we tuned the number of reconstructed  features among $\{500, 1000, 2000, 5000, 10000\}$, the number of model layers among $\{1,3, 5\}$ and the corruption probability 
among $\{0.1, 0.2, \ldots, 0.5\}$. For MSDA-DAN, we followed \newcite{Ganin:16}: the $\lambda$ adaptation parameter is chosen among 9 values between $10^{-2}$ and 1 on a logarithmic scale, the hidden layer size $l$ is chosen among $\{50,100,200\}$ and the learning rate $\mu$ is $10^{-3}$.

\section{Results}

\begin{table*}[t!]
\small
	\centering
	\begin{tabular}{|l|l|l|l|l|l|l|}
		\hline
		Setup  & Gold & Pivots (First doc.)   & Pivots (Second doc.)  & AE-SCL (Fir.,Sec.)  & Rank Diff \\ \hline
		E$\rightarrow$B & 1 & very good,  good &  great    & (1,0) &  58058 (2.90\%) \\ \hline
                E$\rightarrow$D & 1 & fantastic        & wonderful & (1,0) &  44982 (2.25\%) \\ \hline
                K$\rightarrow$E & 1 & excellent, works fine        & well, works well & (1,0)  & 75222 (3.76\%) \\ \hline
                K$\rightarrow$D & 1 & the best,best & perfect	& (1,0) & 98554 (4.93\%) \\ \hline
                D$\rightarrow$B & 0 & boring, waste of  & dull, can't recommend & (1,0) & 78999 (3.95\%) \\ \hline
                B$\rightarrow$D & 0 & very disappointing, disappointing  & disappointed	 & (1,0) & 139851 (6.99\%) \\ \hline
                D$\rightarrow$K & 0 & sadly  & unfortunately & (1,0) & 63567 (3.17\%) \\ \hline
                B$\rightarrow$K & 0 & unhappy   & disappointed & (1,0) & 110544 (5.52\%) \\ \hline
	\end{tabular}
\caption{
Document pair examples from eight setups (1st column)
with the same gold sentiment class. 
In all cases, AE-SCL-SR correctly classifies both documents, while 
AE-SCL misclassifies one (5th column). 
The 6th column presents the difference in the ranking of the cosine scores between the representation vectors $\tilde{h}$ of the documents  
according to both models (the rank of AE-SCL minus the rank of AE-SCL-SR), both  in absolute values and as a percentage  of the 1,999,000 document pairs ($2000 \cdot 1999 /2$) in the test set of each setup. As $\tilde{h}$ is feeded to the sentiment classifer we expect documents that belong to the same class to have more similar $\tilde{h}$ vectors. The differences are indeed positive in all 8 cases.
}
\label{tab:qual}
\end{table*}

\begin{table} [t!]
\small
      \centering
        \begin{tabular}{|l|l|l|}
\hline
           & Positive & Negative \\ 
\hline
AE-SCL-SR & 954 (3.97 \%) & 576 (2.40 \%) \\
\hline
AE-SCL & 527 (2.19 \%) & 754 (3.14 \%) \\
\hline
        \end{tabular}
\smallskip
\\
      \centering
        \begin{tabular}{|l|l|l|}
\hline
       & Positive & Negative \\ 
\hline
AE-SCL-SR  & 1538 (6.40 \%) & 765 (3.18 \%)\\
\hline
MSDA & 673 (2.80 \%) &  1109 (4.60 \%) \\
\hline     
        \end{tabular}
    \caption{
    Class based analysis for the unified test set of the \newcite{Blitzer:07} task. A \textit{(model,class)} presents the number of test examples 
from the \textit{class}, for which the \textit{model} is correct while the other model in the table is wrong.
\vspace{-0.2cm}
} 
\label{tab:per-class-analysis}
\end{table}


Table \ref{tab:results} presents our results. In the \newcite{Blitzer:07} task (top tables), AE-SCL-SR is the best performing model in 9 of 12 setups 
and on a unified test set consisting of the test sets of all 12 setups (the Test-All column).
AE-SCL, MSDA and MSDA-DAN perform best in one setup each. On the unified test set, AE-SCL-SR improves over SCL-MI by 3.8\% (error reduction (ER) of 14.8\%) and over MSDA-DAN by 2\% (ER of 8.4\%), while AE-SCL improves over SCL-MI and MSDA-DAN by 2.7\% (ER of 10.5\%) and 0.9\% (ER of 3.8\%), respectively. MSDA-DAN and MSDA perform very similarly on the unified test set (0.761 and 0.759, respectively) with generally minor differences in the individual setups.

When adapting from the product review domains to the Blog domain (bottom table), AE-SCL-SR performs best in 3 of 4 setups, providing particularly large improvements when training is in the Kitchen (K) domain. The average improvement of AE-SCL-SR over MSDA is 5.2\% and over a non-adapted classifier is 11.7\%. As before, MSDA-DAN performs similarly to MSDA on the unified test set, although the differences in the individual setups are much higher. The differences between AE-SCL-SR and 
the other models
are statistically significant in most cases.\footnote{The difference between two models in a given setup is considered to be statistically significant 
if and only if it is significant in all five folds of that setup.}
 


\paragraph {Class Based Analysis}
Table \ref{tab:per-class-analysis} presents a class-based comparison between model pairs. Results are presented for the unified test set of the \newcite{Blitzer:07} task.
The table reveals that the strength of AE-SCL-SR comes from its improved 
accuracy on positive examples: in 3.97\% of the cases over AE-SCL (compared to 2.19\% of the positive examples where AE-SCL is better) 
and in 6.40\% of the cases over MSDA (compared to 2.80\%).
While on negative examples the pattern is reversed and AE-SCL and MSDA outperform AE-SCL-SR, 
this is a weaker effect which only moderates the overall superiority of AE-SCL-SR.\footnote{The reported numbers are averaged over the 5 folds and rounded to the closest integer, if necessary. The comparison between AE-SCL-SR and MSDA-DAN yields a very similar pattern and is hence excluded from space considerations.} 

The unlabeled documents from all four domains are strongly biased to convey positive opinions
(Section \ref{sec:experiments}). 
This is indicated, for example, by the average score given to these reviews by their authors: 4.29 (B), 
4.33 (D), 3.96 (E) and 4.16 (K), on a scale of 1 to 5.
This analysis suggests that AE-SCL-SR better learns from of its unlabeled data.

\paragraph {Similar Pivots}
Recall that AE-SCL-SR aims to learn more similar representations for documents with similar pivot features. Table \ref{tab:qual} demonstrates this effect through pairs of test documents from 8 product review setups.\footnote{We consider for each setup one example pair from one of the five folds such that the dimensionality of the hidden layers in both models is identical.} 
The documents contain pivot features 
with very similar meaning and indeed they belong to the same sentiment class. 
Yet, in all cases AE-SCL-SR correctly classifies both documents, 
while AE-SCL misclassifies one. 

\com{The rightmost column of the table presents the difference in the ranking of the cosine similarity between the 
representation vectors $\tilde{h}$ of the documents in the pair, according to each of the models. 
For a model $m$ and a document pair $(d1,d2)$ with representation vectors $\tilde{h}^{d1}_m$ and $\tilde{h}^{d2}_m$
we denote $Rank^{d1,d2}_{m} = rank^{m}(cosine(\tilde{h}^{d1}_m, \tilde{h}^{d2}_m))$, where 
$rank^{m}$ is a function that returns the rank of its argument among the cosine similarity values between the representation vectors for any pair of documents $(\tilde{h}^{di},\tilde{h}^{dj})$ according to $m$  (the highest similarity is ranked 1).
The entry in the table hence corresponds to ($Rank^{d1,d2}_{AE-SCL} - Rank^{d1,d2}_{AE-SCL-SR}$), 
presented in numbers and 
as a percentage of the ($2000 \cdot 1999 /2$) document pairs in the test set of each setup. 
As $\tilde{h}$ is fed to the sentiment classifier we expect documents of the same 
class to have more similar $\tilde{h}$ vectors.}

The rightmost column of the table presents the difference in the ranking of the cosine similarity between the 
representation vectors $\tilde{h}$ of the documents in the pair, according to each of the models. Results (in numerical values and percentage) are given with 
respect to all cosine similarity values between the  $\tilde{h}$ vectors  of any document pair in the test set. 
As the documents with the highest similarity are ranked 1, the positive difference between the ranks of AE-SCL and those of AE-SCL-SR indicate that AE-SCL's rank is lower.
That is, AE-SCL-SR learns more similar representations for documents with similar pivot features.
\section{Conclusions and Future Work}


We presented a new model 
for domain adaptation which combines ideas from 
pivot based and autoencoder based representation learning. We have demonstrated how to encode information from pre-trained word embeddings to improve the generalization of our model across examples with semantically similar pivot features. We demonstrated strong performance on cross-domain sentiment classification tasks with 16 domain pairs and provided initial qualitative analysis that supports the intuition behind our model. Our approach is general and applicable for a large number of NLP tasks (for AE-SCL-SR this holds as long as the pivot features can be embedded in a vector space).

In future 
we would like to adapt our model to more general 
domain adaptation setups such as where adaptation is performed between sets of source and target domains and where some labeled data from the target domain(s) is available.


\appendix 
\section{Hyperparameter Tuning}

This appendix describes the hyper-parameter tuning process for the models compared in our paper. Some of these details appear in the full paper, but here we provide a detailed description.

\paragraph {AE-SCL and AE-SCL-SR} 
We tuned the parameters of both our models in two steps. First, we randomly split the unlabeled data 
from both the source and the target domains in a 80/20 manner and combine the large subsets together 
and the small subsets together so that to generate unlabeled training and validation sets. On these training/validation 
sets we tune the hyperparameters of the stochastic gradient descent (SGD) algorithm we employ to train our networks: 
learning rate (0.1), momentum (0.9) and weight-decay regularization ($10^{-5}$). Note that these values are tuned 
on the fully unsupervised task of predicting pivot features occurrence from non-pivot input representation, and are then 
employed in all the source-traget domain combinations, across all folds.
\footnote{Both AE-SCL and AE-SCL-SR converged to the same values. 
This is probably because for each parameter we consider only a handful of values: 
learning rate (0.01,0.1,1), momentum (0.1,0.,5,0.9) and weight-decay regularization ($10^{-4}$,$10^{-5}$, $10^{-6}$).}

After tuning the SGD parameters, in the second step we tuned the model's hyper-parameters for each 
fold of each source-target setup. The hyper-parameters are the
number of pivots (100 to 500 in steps 100) and the dimensionality of $h$ (100 to 500 in steps of 200). 
We select the values that yield the best performing model when training on the training set 
and evaluating on the training domain development set of each fold.\footnote{When tuning the SGD parameters we experimented with 100 and 500 
pivots and dimensionality of 100 and 500 for $h$.} 

We further explored the quality of the various intermediate representations generated by the models as sources of features for the sentiment classifier. The vectors we considered are: $w^{h}x^{np}$, $h = \sigma(w^{h}x^{np})$, $w^{r}h$ and $r = \sigma(w^{r}h)$. We chose the $w^{h}x^{np}$ vector, denoted in the paper in the paper with $\tilde{h}$.

For AE-SCL-SR, embeddings for the unigram and bigram features were 
learned with word2vec \cite{Mikolov:13}.
\footnote{We employed the Gensim  
package and trained the model on the unlabeled 
data from both the source and the target domains of each adaptation setup (\url{https://radimrehurek.com/gensim/}).}
To learn bigram representations, in cases where a bigram pivot \textit{(w1,w2)} is included in a sentence 
we generate the triplet \textit{w1,w1-w2, w2}. For example, the sentence \textit{It was a very good book}  
with the bigram pivot \textit{very good} is re-written as: \textit{It was a very very-good good book}. 
The revised corpus is then fed into word2vec.
The dimension of the hidden layer $h$ of AE-SCL-SR is  the dimension of the induced embeddings.

In both parameter tuning steps we use the unlabeled validation data for early stopping: the SGD algorithm  stops at the first iteration where the validation data error increases rather then when the training error or the loss function are minimized.

\paragraph{SCL-MI}
Following \cite{Blitzer:07} we used 1000 pivot features .\footnote{Results with 500 pivots 
were very similar.} 
The number of SVD dimensions was tuned on the labeled development data to the best value among 
50,100 and 150.

\paragraph{MSDA} Using the labeled dev. data we tuned the number of reconstructed 
features (among 500, 1000, 2000, 5000 and 10000) 
the number of model layers (among $\{1,3, 5\}$) and the corruption probability 
(among $\{0.1, 0.2, \ldots, 0.5\}$). For details on these hyper-parameters see \cite{Chen:12}.

\paragraph{MSDA-DAN} Following \newcite{Ganin:16} we tuned the hyperparameters on the labeled development data as follows. The $\lambda$ adaptation parameter is chosen among 9 values between $10^{-2}$ and 1 on a logarithmic scale. The hidden layer size $l$ is chosen among $\{50,100,200\}$ and the learning rate $\mu$ is fixed to $10^{-3}$.

\section{Experimental Choices}


\paragraph{Variants of the Product Review Data}

There are two releases of the datasets of the \newcite{Blitzer:07} cross-domain product review task. We use the one from \url{http://www.cs.jhu.edu/~mdredze/datasets/sentiment/index2.html} where the data is imbalanced, consisting of more positive than negative reviews. We believe that our setup is more realistic as when collecting unlabeled data, it is hard to get a balanced set. Note that \newcite{Blitzer:07} used the other release where the unlabeled data consists of the same number of positive and negative reviews.

\paragraph{Test Set Size}

While \newcite{Blitzer:07} used only 400 target domain reviews for test, we use the entire set of 2000 reviews. We believe that this decision yields more robust and statistically significant results.

\com{
\paragraph{Blog as a Source Domain}

We do not use the Blog domain as a source domain as it was hard to extract pivot features that are both frequent and provide quality information about the sentiment labels in this domain. This is due to the large number of topics covered in this domain and the individual examples being sentences rather than documents.
}

\bibliography{da}
\bibliographystyle{acl_natbib}

\end{document}